# Compiling Fuzzy Logic Control Rules to Hardware Implementations[†]

Stephen Chiu and Masaki Togai

Rockwell International Science Center
Thousand Oaks, California 91360

## ABSTRACT

*A major aspect of human reasoning involves the use of approximations. Particularly in situations where the decision-making process is under stringent time constraints, decisions are based largely on approximate, qualitative assessments of the situations. Our work is concerned with the application of approximate reasoning to real-time control. Because of the stringent processing speed requirements in such applications, hardware implementations of fuzzy logic inferencing are being pursued. We describe a programming environment for translating fuzzy control rules into hardware realizations. Two methods of hardware realizations are possible. The first is based on a special purpose chip for fuzzy inferencing. The second is based on a simple memory chip. The ability to directly translate a set of decision rules into hardware implementations is expected to make fuzzy control an increasingly practical approach to the control of complex systems.*

## 1. INTRODUCTION

Computer systems that mimick the reasoning of human experts are finding increased applications. A fundamental problem in the design of such systems is that human knowledge cannot always be represented as precise rules. It is also problematic whether we should seek to represent this knowledge as precise rules. Precision is costly in terms of both the amount of information that must be stored and the amount of processing required. The imprecise nature of human rules allows us to make decisions in a timely manner based on approximations.

The cost of using approximation is that the resultant decisions may be sub-optimal. However, since real-world situations tend to be non-deterministic, there is no assurance that any decision will be optimal in its outcome. An optimal decision only exists in the sense of maximizing the probability of a successful outcome. Because the result of any decision cannot be precisely predicted, and that the decision-making process is often under stringent time constraints, real-world decisions are based largely on approximate, qualitative assessments of the situations.

Fuzzy logic is a method of approximate reasoning that deals directly with qualitative decision rules. Expert systems based on fuzzy logic have shown to be effective in a variety of industrial control applications [1]. The acquired experiences of human operators are naturally expressed as fuzzy control rules. The computational efficiency of fuzzy logic inferencing also makes fuzzy control a viable alternative to conventional control methods. Applications have ranged from the control of a cement kiln [2] to a train operation system [3]. A comprehensive survey of fuzzy control is given in [4].

Although the fuzzy inference process is computationally efficient, the stringent time constraints of many systems still preclude the application of fuzzy control. Our work is focused on the implementation of fuzzy control in hardware to achieve the required real-time performance. Two methods of hardware implementations are being pursued. The first is based on a VLSI chip that can perform an entire inference process required for fuzzy control [5,6]. It is currently being designed using $2\mu$ CMOS technology. The developed chip will be capable of processing 16 control rules in parallel, producing 250,000 inferences per second at 16 MHz clock. The second method is based on precomputing the inference conclusions for all possible input states and storing the results in a simple memory chip. During real-time operation, each input state is then directly converted into an addresss for fetching the conclusions.

---

[†] This work was supported by the Independent Research and Development Program of Rockwell International Corporation.

To facilitate the implementation of these control systems, a fuzzy logic programming environment is being developed to translate linguistically expressed rules into these silicon implementations. The environment allows a human expert to easily describe a set of rules, modify them, and verify the rules through simulation. To reflect the final hardware architecture, each set of rules is associated with a chip object simulated in software. The content of the chip object is changed as the rules are modified. When the simulation results are satisfactory, memory maps can be generated from the content of the chip object for writing into either a fuzzy inference chip or a simple memory chip.

In this paper, we describe the programming environment, currently implemented on a Xerox 1186 Lisp workstation. The ability to directly translate a set of decison rules into hardware realizations is expected to make fuzzy control an increasingly practical approach to the control of complex systems.

## 2. FUZZY CONTROL

### 2.1 Rule Representation

The application of fuzzy logic to the control of physical processes was first demonstrated by Mamdani and Assilian [7]. Human strategies for controlling a steam engine was verbally expressed and encoded as fuzzy decision rules. The rules were expressed in the following form:

> IF the pressure error is negative small and
>    the change in pressure error is positive big
> THEN make the heat change positive big.

This rule can be written more concisely as

> IF $X_1$ is NS and $X_2$ is PB
> THEN $Y$ is PB

where $X_1$ and $X_2$ are the inputs to the controller (pressure error and change in pressure error), $Y$ is the output of the controller (heat change), and NS and PB are **fuzzy variables** defining what is meant by negative small and positive big. A fuzzy variable, say $A$, is represented by a distribution function $A(x)$, called the membership function. The NS membership function, for example, determines the level of truth that a given pressure error is negative small. The membership functions are usually chosen as either normal or triangular distributions, as shown in Fig. 1. The range of possible values for the argument, e.g. from -1 to +1, is called the *universe of discourse* of the fuzzy variable.

### 2.2 Fuzzy Reasoning

In fuzzy control, the compositional method of inference is usually used for computational simplicity. Suppose we have the following two rules:

364

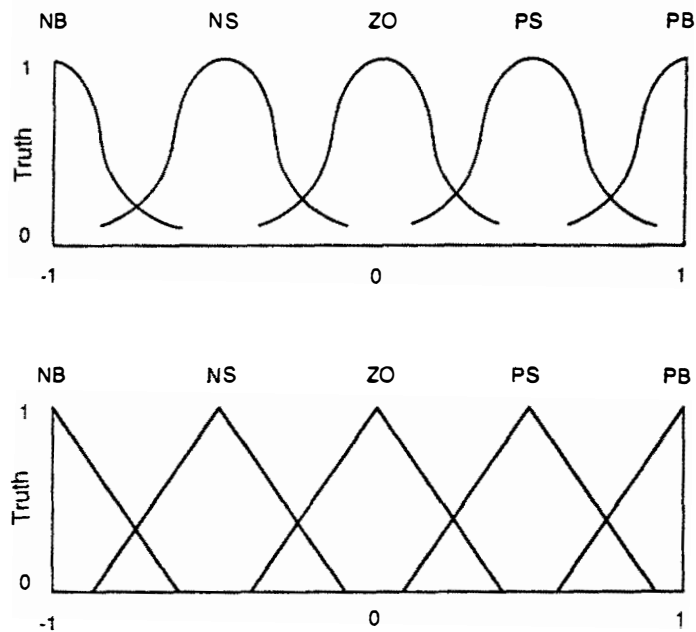

**Figure 1.** Normal and triangular membership functions.

Rule 1:   IF $X_1$ is $A_{11}$ and $X_2$ is $A_{12}$
        THEN $Y$ is $B_1$

Rule 2:   IF $X_1$ is $A_{21}$ and $X_2$ is $A_{22}$
        THEN $Y$ is $B_2$.

Given inputs $x_1$ and $x_2$, the truth values $\alpha_1$ and $\alpha_2$ of the propositions are given by

$$\alpha_1 = A_{11}(x_1) \wedge A_{12}(x_2)$$

$$\alpha_2 = A_{21}(x_1) \wedge A_{22}(x_2)$$

where $\wedge$ denotes the minimum operator.

We form an output membership function $B(y)$ as

$$B(y) = (\alpha_1 \wedge B_1(y)) \vee (\alpha_2 \wedge B_2(y)) \tag{1}$$

where $\vee$ denotes the maximum operator. Often the output membership function is computed as

$$B(y) = (\alpha_1 \times B_1(y)) \vee (\alpha_2 \times B_2(y)). \tag{2}$$

However, the form given by Eq. 1 is more efficient and suited for VLSI implementation, since it only requires min-max comparisons and no multiplications. Hence, it is the method used in the fuzzy inference chip.

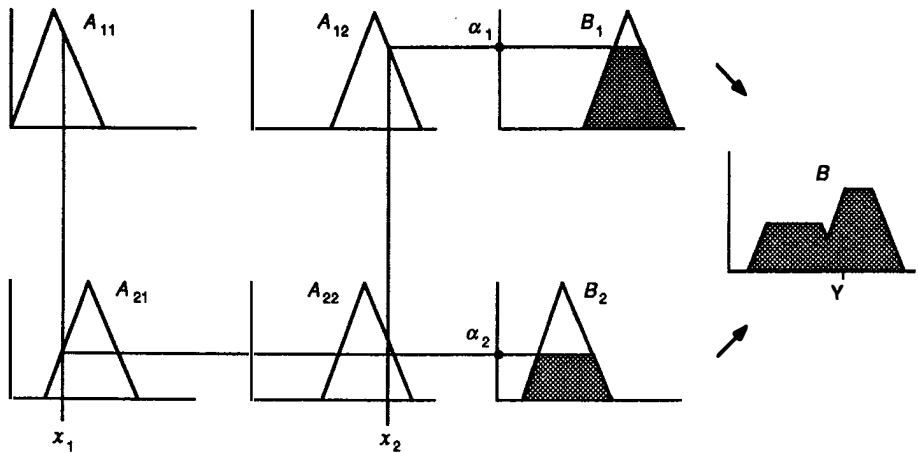

Figure 2. Fuzzy reasoning process utilized by the inference chip.

The output $Y$ is inferred from the centroid of the output membership function:

$$Y = \frac{\int B(y)\, y\, dy}{\int B(y)\, dy}$$

The inference process is illustrated in Fig. 2.

## 3. RULE PROGRAMMING

Controllers are functional subsystems that have clearly defined inputs and outputs. The object-oriented approach of the programming environment explicitly models this functionality. In this environment, each set of rules is associated with a chip object. A set of rules is tested by applying inputs to the corresponding chip object. The rules may be interactively modified based on the observed outputs.

### 3.1 Rule Format

The control rules are defined in text files by using a standard text editor. A simple rule file is shown in Fig. 3. In this example, TEMPERATURE and PRESSURE are the inputs to the controller; HEATER.POWER and VALVE.OPENING are the outputs. The numerical expressions, e.g. (0 200), define the universe of discourse for each input and output. For example, the expression (0 200) defines the input temperature range to be 0 to 200 degrees. Adjectives such as HIGH.TEMP and LOW.PRESS are fuzzy variables that are each associated with a membership function. The adverbs ABOVE and VERY modify the membership functions, as will be described later.

Because the fuzzy inference chip is designed to process conjunctive rules, i.e. those consisting exclusively of AND logic, the rule file is currently restricted to contain only conjunctive rules. However, a disjunctive rule can always be formulated in terms of a set of conjunctive rules. For example, the rule

    IF   ($X_1$ is NS and $X_2$ is PB) or ($X_1$ is NB)
    THEN  $Y$ is PB

```
(* Rules for Boiler Control)

(INPUT  TEMPERATURE (0 200)  PRESSURE (0 500))
(OUTPUT HEATER.POWER (0 10)  VALVE.OPENING  (0 10))

(IF TEMPERATURE IS HIGH.TEMP AND
    PRESSURE IS LOW.PRESS
 THEN
     HEATER.POWER IS LOW AND
     VALVE.OPENING IS SMALL)

(IF TEMPERATURE IS ABOVE AVERAGE.TEMP AND
    PRESSURE IS VERY HIGH.PRESS
 THEN
    HEATER.POWER IS LOW AND
    VALVE.OPENING IS VERY LARGE)
```

**Figure 3.** Rule file format.

is equivalent to the two conjunctive rules:

Rule 1:  IF  $X_1$ is NS and $X_2$ is PB
         THEN $Y$ is PB

Rule 2:  IF  $X_1$ is NB
         THEN $Y$ is PB.

It is possible to write a rule compiler that will translate disjunctive rules into equivalent conjunctive rules.

### 3.2 Defining Fuzzy Variables

The fuzzy variables are defined by using a graphic editor (Fig. 4). The discretized membership function corresponds to the representation in the fuzzy inference chip; the universe of discourse is discretized into 16 levels of resolution, each with 16 levels of truth value. The discretization can be easily changed to adapt the environment to new types of chips.

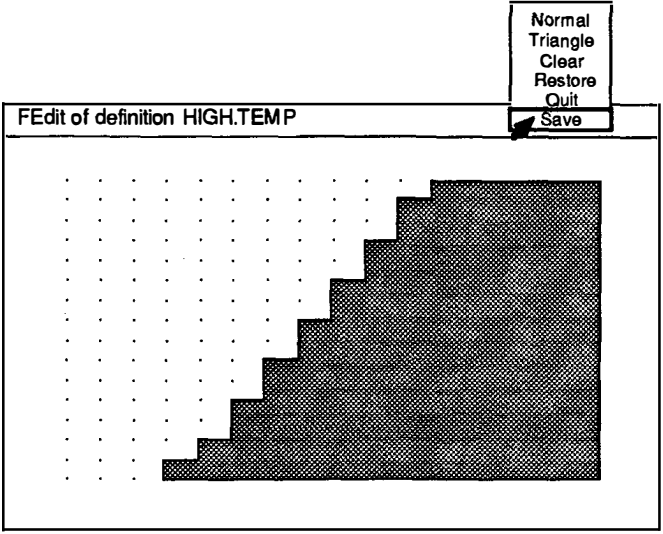

**Figure 4.** A fuzzy definition editor window.

367

Definition editing is performed via the mouse. A user can define normal or triangular distributions by selecting from a options menu. After selecting either the "Normal" or the "Triangle" option, the first mouse click on grid marks the center of the distribution and the second click marks the tail of the distribution. Arbitrary distributions can also be defined by simply drawing with the mouse. By selecting the "Save" option from the menu, the definition is saved in a dictionary of fuzzy variables.

Adverbs may be used in rules to modify the definition of a fuzzy variable. Some adverbs, such as "above" and "below", are used to indicate a relation with a definition. Others, such as "very" and "somewhat", are used to narrow or relax a definition. The effects of some adverbs are shown in Fig. 5.

## 3.3 The Chip Object

Chip objects are convenient abstractions for performing translation and verification of rules. Fig. 6 shows the sequence of user commands for translating and verifying a set of rules, which are readily expressed as operations on a chip object. In this example, the user first creates a chip object named MYCHIP. The rule file BOILER.RULE is then read into the chip object. The rule parser interprets the text and assembles the appropriate numerical membership functions into a data structure stored in MYCHIP. The rules can then be tested by applying inputs to the chip object. The output is returned as the result of applying input. The output membership functions may also be examined graphically. If the output is not satisfactory, the definition editor can be used to adjust the definition of the fuzzy variables. The content of the chip object is automatically updated to reflect any change in definitions. When the simulation results are satisfactory, the content of the chip object then represents the state to be duplicated in the hardware chip.

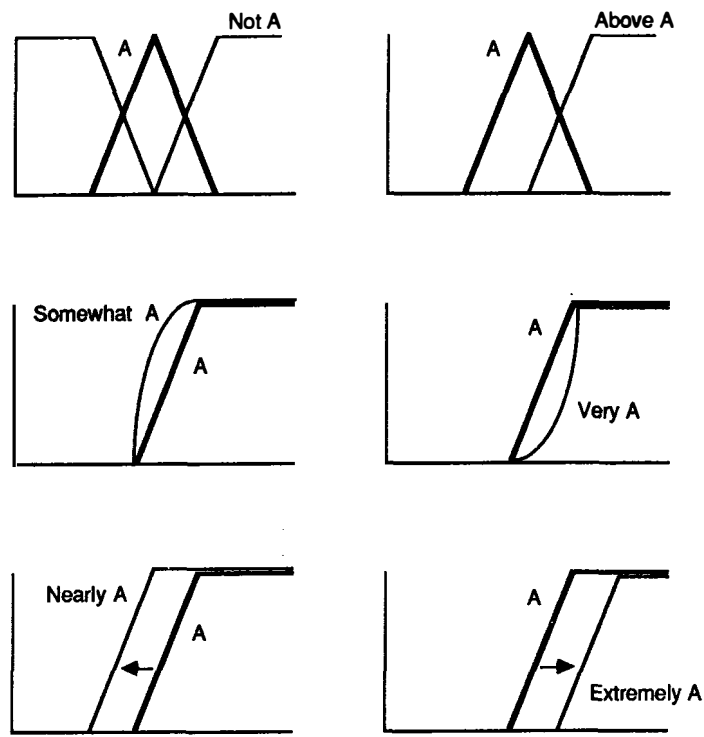

**Figure 5.** Some adverb modifiers.

```
(SETQ MYCHIP (CREATE-OBJECT CHIP :TYPE MINMAX) )  ----> sets MYCHIP to
                                                       be a chip object

(READRULE 'BOILER.RULE MYCHIP)  ----> reads a rule file into MYCHIP

(ASSERT-INPUT '(150 200) MYCHIP) => (4.35 2.82)

(ASSERT-INPUT '(150 400) MYCHIP) => (2.91 8.87)

(DISPLAY-OUTPUT MYCHIP) -----> displays the output membership functions
```

**Figure 6.** Sequence of user commands for verifying the rules.
The symbol "=>" indicates the value returned by a function.

The method of fuzzy reasoning applied to the rules is specified by creating chip objects of the appropriate type. Currently two types of chip objects are defined. Chip objects of the "Minmax" type utilize the min-max operation to combine rules (Eq. 1), and chip objects of the "Multiplicative" type utilize the max-product operation (Eq. 2). Although chip objects to be implemented as fuzzy inference chips must be of the Minmax type, there is no restriction on the type of chip objects that can be implemented as memory chips.

By using a CONNECT function, the chip objects may be interconnected in such a way that the output of a chip becomes the input to any number of other chips. When an input is asserted on a chip, its output is propagated to all its "downstream" chips. Hence, a circuit of cascaded chips may be easily simulated.

## 4. HARDWARE TRANSLATION

At the current stage of development, files can be generated from the content of chip objects that correspond to memory maps for the hardware chips. Direct translation to hardware will be accomplished in the future when physical links are established between the programming workstation and devices for writing into chip memory locations.

### 4.1 Inference Chip Format

The fuzzy inference chip utilizes min-max circuits to process membership functions stored in its memory unit [6]. It is designed to process rules with 4 input and 2 output variables, i.e. rules of the form

IF $X_1$ is $A_1$ and $X_2$ is $A_2$ and $X_3$ is $A_3$ and $X_4$ is $A_4$
THEN $Y_1$ is $B_1$ and $Y_2$ is $B_2$ .

Because mechanized hardware processing requires all rules to conform to the 4-input/2-output format, existing rules usually need to be padded with "dummy" membership functions. For example, the rule

IF $X_1$ is NS and $X_2$ is PB
THEN $Y_1$ is PB

is translated into

IF $X_1$ is NS and $X_2$ is PB and $X_3$ is ANY and $X_4$ is ANY
THEN $Y_1$ is PB and $Y_2$ is NULL

where ANY is a membership function of all maximum truth values and NULL is a membership function of all zero truth values.

By applying a WRITERULE function to a chip object, a data structure that conforms to the hardware chip representation of the rules is generated and written into a file. The file basically contains sequences of membership functions, represented as vectors of 4-bit integer truth values. When the physical link between the workstation and the write device for the inference chip is established, the memory unit on a chip will be directly accessed for writing.

### 4.2 Memory Chip Format

If a fuzzy inference chip has only a few inputs, the function of the inference chip can be efficiently implemented by a simple memory chip. Consider a 2-input inference chip, where each input is a 4-bit integer; the entire input state to the chip is represented by 8 bits. The chip effectively functions by using the 8-bit input state to generate a set of output values. In this way, the inference chip is similar to a simple memory unit, i.e. an 8-bit "address" is used to "fetch" a set of output values. This similarity suggests the alternate approach to implementing fuzzy control in hardware. A fuzzy controller can be based on a memory chip by using the input state as an address and storing the precomputed output values at the corresponding address location.

To implement a chip object as a memory chip, a GENTABLE function is used to generate an address table. Figure 7 shows a typical address table generated from a 2-input/2-output chip object. The first column is an integer representing the address. The columns following the address show the individual input values which make up the address. The individual input values are useful as array indices when implementing the control functions as an array lookup table in a computer program, as suggested in [8]. The columns following the input values are the corresponding output values. In this example, the output values are represented as floating point numbers. Alternatively, a user may specify the output to be represented as an integer of a specified size. For example, if the output size is specified as 8 bits, the output values will be represented as integers between 0 and 255. In practical applications, the output representation is usually chosen based on memory size limitations or D/A converter resolution.

```
INPUT 2   OUTPUT 2   BYTESIZE 0

0         0          0          12.46835    2.493976
1         1          0          12.46835    2.493976
2         2          0          12.35135    2.602564
3         3          0          12.20896    2.728572
4         4          0          12.0339     2.85
.         .          .          .           .
.         .          .          .           .
15        15         0          3.266667    2.0
16        0          1          2.709678    2.0
17        1          1          2.204082    1.8
18        2          1          1.916667    1.592593
.         .          .          .           .
.         .          .          .           .
255       15         15         2.582279    8.043165
```

**Figure 7.** Address table for memory chip.

## 5. CONCLUSION

A programming environment for translating fuzzy control rules to hardware implementations has been described. In this environment, chip objects provide a convenient abstraction for the development of control rules. An abstract chip object may readily be translated into a physical realization as either a fuzzy inference chip or a simple memory chip. The inference chip realization is suited for applications in which the number of possible input states is large, and in which the rules may need to be changed dynamically. The memory chip realization is suited for applications in which the the number of possible input states is small and the rules are constant. These hardware implementations of fuzzy logic constitute a step in making approximate reasoning a practical component in real-time systems.

## REFERENCES

[1] M. Sugeno (editor), *Industrial Applications of Fuzzy Control*, North-Holland, Amsterdam, 1985.

[2] L.P. Holmblad and J.J. Ostergaard, "Control of a cement kiln by fuzzy logic," in *Fuzzy Information and Decision Processes* (M.M. Gupta and E. Sanchez, eds.), North-Holland, Amsterdam, 1982.

[3] M. Miyamoto and S. Yasunobu, "Predictive fuzzy control and its application to automatic train operation systems," *Proc. First Int. Conf. on Fuzzy Information Processing*, Hawaii, July 1984.

[4] M. Sugeno, "An introductory survey of fuzzy control," *Information Sciences*, Vol. 36, 1985, pp. 59-83.

[5] M. Togai and H. Watanabe, "A VLSI implementation of a fuzzy-inference engine: toward an expert system on a chip," *Information Sciences*, Vol. 38, 1986, pp. 147-163.

[6] M. Togai and S. Chiu, "A fuzzy logic accelerator and a programming environment for real-time fuzzy control," *Proc. 2nd Int. Fuzzy Systems Assoc. Congress*, Tokyo, Japan, July 1987.

[7] E.H. Mamdani and S. Assilian, "A case study on the application of fuzzy set theory to automatic control," *Proc. IFAC Stochastic Control Symposium*, Budapest, 1974.

[8] D.A. Rutherford and G.C. Bloore, "The implementation of fuzzy algorithms for control," *Proceedings of the IEEE,* Proceedings Letters, April 1976, pp. 572-573.